# Novel Co-variant Feature Point Matching Based on Gaussian Mixture Model

Liang Shen, Jiahua Zhu, Chongyi Fan, Xiaotao Huang, *Member, IEEE* and Tian Jin

*Abstract*—The feature frame is a key idea of feature matching problem between two images. However, most of the traditional matching methods only simply employ the spatial location information (the coordinates), which ignores the shape and orientation information of the local feature. Such additional information can be obtained along with coordinates using general co-variant detectors such as DOG, Hessian, Harris-Affine and MSER. In this paper, we develop a novel method considering all the feature center position coordinates, the local feature shape and orientation information based on Gaussian Mixture Model for co-variant feature matching. We proposed three sub-versions in our method for solving the matching problem in different conditions: rigid, affine and non-rigid, respectively, which all optimized by expectation maximization algorithm. Due to the effective utilization of the additional shape and orientation information, the proposed model can significantly improve the performance in terms of convergence speed and recall. Besides, it is more robust to the outliers.

## I. Introduction

FEATURE point sets matching problem is defined as the establishment of correct correspondence between two feature sets [1]. It is an important problem in pattern recognition, computer vision and remote sensing, which often used for object recognition [2], image registration [1, 3], stereo correspondence [4] and so on. The problem can be divided into two parts: feature extraction and feature matching.

Feature extraction is operated using the feature detectors. These detectors may be Morvec [5], Harris, Fast, Harris-affine[6], Scale-Invariant Fourier Transform (SIFT) [7], Salient regions [8], MSER [9], ASV[10] and so on. Among the various descriptors, the co-variant feature detectors such as SIFT, Harris–affine & Hessian-affine, Salient regions [8] and MSER [9] plays a fundamental role, which are widely used for different applications [1, 3, 11, 12]. The goal of feature matching is to find the correspondence between two feature point sets. Many methods have been proposed for this problem [13-18]**.**

The simplest way is finding the correspondence using similarity constraint, which refers to that points only match points with the most similar local descriptors. Popular methods of this kind include SIFT [7] and shape context [13]. These methods typically obtain not only most of the correct matches (i.e. inliers) but also many false matches (i.e. outliers) due to the ambiguity of similarity constraint. To overcome the problem, using geometrical constraints to remove the outliers is a popular way, which involving that matches should satisfy an underlying geometrical requirement. The classical RANSAC [19] algorithm is one such example. Other examples include graph shift [2], vector field consensus [20], and Locally Linear Transforming [1]. The geometric transformational models used in these methods include both parametric, such as rigid and affine [21, 22], and nonparametric models such as thin-plate spline [23]. In addition, formulating the matching problem in terms of a correspondence matrix between the control points based on the geometrical constraint is also a popular way. Examples of this strategy include the well-known iterative closest point algorithm [24], Chui and Rangarajan's method [25].

Among the various methods, the Gaussian Mixture Model (GMM) [1, 16-18, 26-28] has got great attention. GMM based methods treat each point in one set as the GMM centroid and the points in the other set as the data points. The two sets are, accordingly call the model set and data set. Typically, the model set is forced to move towards the data set using a certain transformation function (such as rigid and affine) until they are spatially aligned. A unified framework to both the rigid and non-rigid matching problem is proposed in [18], which uses the L2 distance to measure the similarity between Gaussian mixtures and can reinterpret many other methods [16, 27]

Liang Shen, is with College of Electronic Science, National University of Defense Technology, Changsha, 410000, China (e-mail: shenliang16@nudt.edu.cn ).
Jiahua Zhu, is with College of Meteorology and Oceanology, National University of Defense Technology, Changsha, 410000, China (e-mail: Zhujiahua1019@hotmail.com ).
Chongyi Fan, is with College of Electronic Science, National University of Defense Technology, Changsha, 410000, China (e-mail: Chongyifan@sina.com )
Xiaotao Huang is with College of Electronic Science, National University of Defense Technology, Changsha, 410000, China (e-mail: xtHuang@nudt.edu.cn ) meaningfully. The famous coherence point drift (CPD) algorithm [16] also provides a good framework for point set registration. It includes an additional uniform distribution in the mixture model to account for outliers and noise and is optimized by expectation maximization. Besides, authors in [26] developed a registration method using particle filtering and stochastic dynamics. The above mentioned GMM-based methods all use the uniform GMM framework, where all the Gaussian components are given the same weights. In such framework, only the position information of the point sets is considered. The authors proposed Non-uniform GMM (NGMM) for feature point sets matching in [17] to Fuse descriptor and Spatial Information. It improves the performance in both finding matches and discarding mismatches on high-quality optical datasets.

In this paper, we notice the characteristic that the co-variant detectors can provide the shape and orientation information of the feature point (most of the traditional methods only uses the feature center location information). In this paper, we propose a GMM based method employing all the information. The shape and orientation information are represented in an affine adaptation matrix, and the complexity of our model is well-controlled under a proper independence assumption. Experiments demonstrates that the efficiency and robustness can both be enhanced using our method as more information are used. To the best of our knowledge, it is the first time for us to integrate the shape information into the GMM-based point sets registration framework; and propose a novel feature frame matching scheme.

## II. BACKGROUND

In this section, we introduce two foundations of this paper: 1) the basic mathematical model of GMM based matching, 2) the co-variant features.

### A. Location-based GMM Matching

In GMM based matching model, the alignment of two point sets is considered as a probability density estimation problem, where one point set represents the Gaussian mixture model (GMM) centroids, and the other one represents the data points. At the optimum, two point sets become aligned and the correspondence is obtained using the maximum of the GMM posterior probability for a given data point. Core to the method is to force GMM centroids to move coherently as a group to preserve the topological structure of the point sets.

Given feature set $\mathbf{X}_{D\times N}=(\mathbf{x}_1,...,\mathbf{x}_N)$ as $N$ data points and set $\mathbf{Y}_{D\times M}=(\mathbf{y}_1,...,\mathbf{y}_M)$ as $M$ GMM centroids. Each point or centroid represents the feature location (i.e. coordinates). $D$ is the dimension of the coordinates. The Transformation from $\mathbf{Y}$ to $\mathbf{X}$ is denoted as $\mathcal{T}(\mathbf{Y},\theta)$, where $\theta$ is a set of transformation parameters. In addition, we use $\mathbf{I}$ as the identity matrix, $\mathbf{1}$ as the column vector of all ones, and d($\mathbf{a}$) as the diagonal matrix formed from the vector $\mathbf{a}$ throughout the paper. Generally, the GMM (with outliers and noise) takes the form

$$p(\mathbf{x})=\omega\frac{1}{N}+(1-\omega)\sum_{m=1}^{M}\frac{1}{M}p(\mathbf{x}|m) \quad (1)$$

where $p(\mathbf{x}|m)=\frac{1}{(2\pi)^{D/2}|\mathbf{\Sigma}|}\exp^{(\mathbf{x}_n-\mathcal{T}(\mathbf{y}_m,\theta))^T\mathbf{\Sigma}^{-1}(\mathbf{x}_n-\mathcal{T}(\mathbf{y}_m,\theta))}$ is the $m^{\text{th}}$ Gauss distribution of $\mathbf{x}_n$, $\mathbf{\Sigma}$ is the covariance matrix (generally, dimensions of coordinates are assumed to be independent of each other and have the same Gauss bandwidths). The additional uniform distribution $\frac{1}{N}$ accounts for outliers and noise, where the weight is $\omega$ ($0<\omega<1$). And all GMM components share the same probability $\frac{1}{M}$. Then the negative log-likelihood function is written as

$$E(\theta,\mathbf{\Sigma})=-\sum_{n=1}^{N}\log\sum_{m=1}^{M+1}p(m)p(\mathbf{x}|m) \quad (2)$$

Generally, Expectation Maximization (EM) algorithm [29, 30] is used for optimization and parameters estimation. We firstly give the complete negative log-likelihood (or called $Q$ function):

$$Q=-\sum_{n=1}^{N}\sum_{m=1}^{M+1}p_{mn}\log(\omega\frac{1}{N}+(1-\omega)\sum_{m=1}^{M}\frac{1}{M}p(\mathbf{x}|m)) \quad (3)$$

where $p_{mn}$ is the posteriori probability distribution of a mixture component. With $N_p=\mathbf{1}^T\mathbf{P}\mathbf{1}$ and ignorance of the constants independent of $\theta$ and $\mathbf{\Sigma}$, we rewrite (3) as

$$\begin{aligned}Q(\theta,\mathbf{\Sigma})=&\frac{1}{2}\sum_{n,m=1}^{N,M}p_{mn}(\mathbf{x}_n-\mathcal{T}(\mathbf{y}_m,\theta))^T\mathbf{\Sigma}^{-1}(\mathbf{x}_n-\mathcal{T}(\mathbf{y}_m,\theta))\\&+N_p\log|\mathbf{\Sigma}|+\frac{N_pD}{2}\log(2\pi)-N_p\log(\frac{1-\omega}{M})\\&-(N-N_p)\log(\frac{\omega}{N})\end{aligned} \quad (4)$$

The EM algorithm proceeds by alternating between E- and M-steps until convergence. E-step: Estimating the probability $p_{mn}$ using (5), which indicates what degree the correspondence ($\mathbf{x}_m-\mathbf{y}_n$) belonging to the inlier set under the given parameter set $\theta$. And in the M-step, the parameters $\theta$ and $\mathbf{\Sigma}$ are updated based on the estimated responsibility.

$$p_{mn}=\frac{e^{-\frac{1}{2}(\mathbf{x}_n-\mathcal{T}(\mathbf{y}_m,\theta))^T\mathbf{\Sigma}^{-1}(\mathbf{x}_n-\mathcal{T}(\mathbf{y}_m,\theta))}}{\sum_{m=1}^{M}e^{-\frac{1}{2}(\mathbf{x}_n-\mathcal{T}(\mathbf{y}_m,\theta))^T\mathbf{\Sigma}^{-1}(\mathbf{x}_n-\mathcal{T}(\mathbf{y}_m,\theta))}+(2\pi)^{\frac{D}{2}}|\mathbf{\Sigma}|\frac{\omega}{1-\omega}\frac{M}{N}} \quad (5)$$

### B. Co-variant Feature Frame

In this part, we discuss about the shape, scale and orientation information. In detecting co-variant features, the location, shape and orientation of a feature $\mathcal{X}_n$ can be defined by an *affine adaptation matrix* [6, 31, 32] as (6) shows, where the location $\mathbf{x}_n\in\square^{2\times 1}$ represents the translation and $\mathbf{A}_n\in\square^{2\times 2}$ is a linear map representing the shape, scale and orientation information.

$$\mathcal{X}_n = \begin{vmatrix} \mathbf{A}_n & \mathbf{x}_n \\ \mathbf{0} & 1 \end{vmatrix} = \begin{vmatrix} \mathcal{X}_{11} & \mathcal{X}_{12} & \mathcal{X}_{13} \\ \mathcal{X}_{21} & \mathcal{X}_{22} & \mathcal{X}_{23} \\ 0 & 0 & 1 \end{vmatrix} \quad (6)$$

The affine matrix denotes the shape of an image region, which constructs an affinely co-variant feature frame. Such frame can be obtained using Mikolajczyk's method [6] or directly using [32]. In such case, the geometry configuration of a feature can be regarded as a six-dimensional point.

The purpose of a frame is twofold. First, it defines a local image region, which denotes where the feature are. Second, it also specifies a transformation. Specifically, a feature pair $\mathcal{X}_n$ and $\mathcal{Y}_m$ specifies the relative affine transformation $\mathcal{T}_{mn}$:

$$\mathcal{T}_{mn} = \mathcal{X}_n \times \mathcal{Y}_m^{-1} = \begin{vmatrix} \mathbf{B}_{mn} & \mathbf{t}_{mn} \\ \mathbf{0} & 1 \end{vmatrix} = \begin{vmatrix} \mathcal{T}_{11} & \mathcal{T}_{12} & \mathcal{T}_{13} \\ \mathcal{T}_{21} & \mathcal{T}_{22} & \mathcal{T}_{23} \\ 0 & 0 & 1 \end{vmatrix} \quad (7)$$

where $\mathbf{B}_{mn} \in \Box^{2 \times 2}$ is a non-singular matrix indicating the relative transformation of rotation, scaling and shearing, while $\mathbf{t}_{mn} \in \Box^{2 \times 1}$ denotes the relative transformation of translation.

### III. THE PROPOSED LOCAL FEATURE FRAME GAUSS MIXTURE MODELS

The main difference between the normal GMM based matching algorithm and the proposed method is that we integrated the shape and orientation information of the feature point into the matching model.

As aforementioned, the shape and orientation information are all integrated in the affine matrix of the feature frame, which has 6 parameters. Therefore, the 2-dimentional location matching problem becomes to a 6-dimentional ($D = 6$) affine matrix matching. Firstly, we define the features $\mathbf{x}_n$ and $\mathbf{y}_m$ according the vector form of their affine matrix:

$$\mathbf{x}_n = [\mathcal{X}_{11}^n, \mathcal{X}_{21}^n, \mathcal{X}_{12}^n, \mathcal{X}_{22}^n, \mathcal{X}_{13}^n, \mathcal{X}_{23}^n]^T$$
$$\mathbf{y}_m = [\mathcal{Y}_{11}^m, \mathcal{Y}_{21}^m, \mathcal{Y}_{12}^m, \mathcal{Y}_{22}^m, \mathcal{Y}_{13}^m, \mathcal{Y}_{23}^m]^T$$

Then, (7) can be rewritten as:

$$\mathbf{x}_n = \mathcal{B}_{mn} \times \mathbf{y}_m + \mathbf{t} \quad (8)$$

where $\mathcal{B}_{mn} = d(\mathbf{B}_{mn}, \mathbf{B}_{mn}, \mathbf{B}_{mn}) \in \Box^{6 \times 6}$ is a square diagonal matrix indicating the 6-dimentional transformation between $\mathbf{x}_n$ and $\mathbf{y}_m$, $\mathbf{t}^{6 \times 1} = \begin{vmatrix} \mathbf{0}^{1 \times 4} & \mathbf{t}_{mn}^T \end{vmatrix}^T$ is the translation vector.

#### A. Rigid Feature Matching

In the global rigid transformation model, all the members share the same relative transformation. And specially, the transformation only involves rotation and translation, which can be defined as the following form:

$$\mathcal{T} = \begin{vmatrix} \mathbf{B} & \mathbf{t} \\ \mathbf{0} & 1 \end{vmatrix} = \begin{vmatrix} s\mathbf{r} & \mathbf{t} \\ \mathbf{0} & 1 \end{vmatrix}$$

where $\mathbf{r}^{2 \times 2}$ is the rotation matrix. Then (8) we obtain $\mathcal{B} = sd(\mathbf{r},\mathbf{r},\mathbf{r})$ and we define the transformation as $\mathcal{T}(\mathbf{y}_m) = \mathcal{B}\mathbf{y}_m + \mathbf{t}$, where $\mathbf{t}^{6 \times 1} = [0,0,0,0,t_1,t_2]^T$ is the translation vector, $\mathbf{B} = d(\mathbf{r}^{2\times2}, \mathbf{r}^{2\times2}, \mathbf{r}^{2\times2})$ (), and $s$ is a scaling parameter. Then the objective function $Q$ becomes:

$$Q(\mathcal{B}, \mathbf{t}, \Sigma) = \frac{1}{2} \sum_{n,m=1}^{N,M} p_{mn} [(\mathbf{x}_n - \mathcal{B}\mathbf{y}_m - \mathbf{t})^T \Sigma^{-1} (\mathbf{x}_n - \mathcal{B}\mathbf{y}_m - \mathbf{t})]$$
$$+ N_p \log(|\Sigma|) + \frac{N_p D}{2} \log(2\pi) - N_p \log(1-\omega) \quad (9)$$
$$- (N - N_p) \log(\omega), \quad \text{s.t. } \mathbf{R}^T \mathbf{R} = \mathbf{I}, \det(\mathbf{R}) = 1$$

where $N_p = \mathbf{1}^T \mathbf{P} \mathbf{1}$. To simplify problem (9), we represent $\mathbf{x}_n$ and $\mathbf{y}_m$ with:

$$\mathbf{x}_n = \begin{vmatrix} \dot{\mathbf{x}}_n \\ \ddot{\mathbf{x}}_n \\ \dddot{\mathbf{x}}_n \end{vmatrix}, \quad \mathbf{y}_m = \begin{vmatrix} \dot{\mathbf{y}}_m \\ \ddot{\mathbf{y}}_m \\ \dddot{\mathbf{y}}_m \end{vmatrix}, \text{ where }$$

$$\dot{\mathbf{x}}_n = [\mathcal{X}_{11}^n, \mathcal{X}_{21}^n]^T, \quad \ddot{\mathbf{x}}_n = [\mathcal{X}_{12}^n, \mathcal{X}_{22}^n]^T, \quad \dddot{\mathbf{x}}_n = [\mathcal{X}_{13}^n, \mathcal{X}_{23}^n]^T,$$
$$\dot{\mathbf{y}}_m = [\mathcal{Y}_{11}^m, \mathcal{Y}_{21}^m]^T, \quad \ddot{\mathbf{y}}_m = [\mathcal{Y}_{12}^m, \mathcal{Y}_{22}^m]^T, \quad \dddot{\mathbf{y}}_m = [\mathcal{Y}_{13}^m, \mathcal{Y}_{23}^m]^T,$$

Obviously, $\dddot{\mathbf{x}}_n = [\mathcal{X}_{13}^n, \mathcal{X}_{23}^n]^T$ and $\dddot{\mathbf{y}}_m = [\mathcal{Y}_{13}^m, \mathcal{Y}_{23}^m]^T$ are the original 2-D coordinates. And, the covariance matrix $\Sigma$ can be denoted as $\Sigma^{6 \times 6} = d(\dot{\sigma}\mathbf{I}^{2\times2}, \ddot{\sigma}\mathbf{I}^{2\times2}, \dddot{\sigma}\mathbf{I}^{2\times2})$. Then

$$(\mathbf{x}_n - \mathcal{B}\mathbf{y}_m - \mathbf{t})^T \Sigma^{-1} (\mathbf{x}_n - \mathcal{B}\mathbf{y}_m - \mathbf{t}) =$$
$$\left\| \frac{\dot{\mathbf{x}}_n - s\mathbf{r}\dot{\mathbf{y}}_m}{\dot{\sigma}} \right\|^2 + \left\| \frac{\ddot{\mathbf{x}}_n - s\mathbf{r}\ddot{\mathbf{y}}_m}{\ddot{\sigma}} \right\|^2 + \left\| \frac{\dddot{\mathbf{x}}_n - s\mathbf{r}\dddot{\mathbf{y}}_m - \mathbf{t}^{2 \times 1}}{\dddot{\sigma}} \right\|^2 \quad (10)$$

Therefore, (9) can be rewritten as

$$Q(\mathbf{r}, \mathbf{t}, s, \Sigma) = \frac{N_p D}{2} \log(2\pi) + 2N_p \log(\dot{\sigma}\ddot{\sigma}\dddot{\sigma})$$
$$+ \frac{1}{2} \sum_{n,m=1}^{N,M} p_{mn} \left( \left\| \frac{\dot{\mathbf{x}}_n - s\mathbf{r}\dot{\mathbf{y}}_m}{\dot{\sigma}} \right\|^2 + \left\| \frac{\ddot{\mathbf{x}}_n - s\mathbf{r}\ddot{\mathbf{y}}_m}{\ddot{\sigma}} \right\|^2 + \left\| \frac{\dddot{\mathbf{x}}_n - s\mathbf{r}\dddot{\mathbf{y}}_m - \mathbf{t}^{2 \times 1}}{\dddot{\sigma}} \right\|^2 \right) \quad (11)$$
$$- N_p \log(1-\omega) - (N - N_p) \log(\omega), \quad \text{s.t. } \mathbf{r}^T \mathbf{r} = \mathbf{I}, \det(\mathbf{r}) = 1$$

To solve (11), first, we eliminate translation $\mathbf{t}$ from $Q$. Taking partial derivative of $Q$ with respect to $\mathbf{t}$ and equate it to zero, we obtain:

$$\mathbf{t} = \frac{1}{N_P} \dddot{\mathbf{X}} \mathbf{P}^T \mathbf{1} - s\mathbf{r} \frac{1}{N_P} \dddot{\mathbf{Y}} \mathbf{P} \mathbf{1}$$
$$= \dddot{\boldsymbol{\mu}}_x - s\mathbf{r}\dddot{\boldsymbol{\mu}}_y$$

where the matrix $\mathbf{P}$ has elements $P_{mn}$ in (6) and the mean vectors $\dddot{\boldsymbol{\mu}}_y$ and $\dddot{\boldsymbol{\mu}}_x$ are defined as:

$$\ddot{\boldsymbol{\mu}}_x = \mathbf{E}(\ddot{\mathbf{X}}) = \frac{1}{N_P}\ddot{\mathbf{X}}\mathbf{P}^T\mathbf{1}, \quad \ddot{\boldsymbol{\mu}}_y = \mathbf{E}(\ddot{\mathbf{Y}}) = \frac{1}{N_P}\ddot{\mathbf{Y}}\mathbf{P}\mathbf{1}.$$

Then, we discuss the solution of $\mathbf{r}$. By substituting $\mathbf{t}$ back into the objective function and rewriting it in matrix form, we get

$$\begin{aligned}Q = \frac{1}{2}\mathrm{tr}[&\dot{\mathbf{X}}'d(\mathbf{P}^T\mathbf{1})\dot{\mathbf{X}}'^T - 2s\dot{\mathbf{X}}'\mathbf{P}^T\dot{\mathbf{Y}}'^T\mathbf{r}^T + s^2\dot{\mathbf{Y}}'d(\mathbf{P}\mathbf{1})\dot{\mathbf{Y}}'^T\\
&+ \ddot{\mathbf{X}}'d(\mathbf{P}^T\mathbf{1})\ddot{\mathbf{X}}'^T - 2s\ddot{\mathbf{X}}'\mathbf{P}^T\ddot{\mathbf{Y}}'^T\mathbf{r}^T + s^2\ddot{\mathbf{Y}}'d(\mathbf{P}\mathbf{1})\ddot{\mathbf{Y}}'^T\\
&+ \dddot{\mathbf{X}}'d(\mathbf{P}^T\mathbf{1})\dddot{\mathbf{X}}'^T - 2s\dddot{\mathbf{X}}'\mathbf{P}^T\dddot{\mathbf{Y}}'^T\mathbf{r}^T + s^2\dddot{\mathbf{Y}}'d(\mathbf{P}\mathbf{1})\dddot{\mathbf{Y}}'^T]\\
&+ 2N_p\log(\dot{\sigma}\ddot{\sigma}\dddot{\sigma})\end{aligned} \quad (12)$$

Where

$$\dot{\mathbf{X}}' = \frac{\dot{\mathbf{X}}}{\dot{\sigma}}, \quad \ddot{\mathbf{X}}' = \frac{\ddot{\mathbf{X}}}{\ddot{\sigma}}, \quad \dddot{\mathbf{X}}' = \frac{\dddot{\mathbf{X}}' - \ddot{\boldsymbol{\mu}}'_x\mathbf{1}^T}{\dddot{\sigma}},$$

$$\dot{\mathbf{Y}}' = \frac{\dot{\mathbf{Y}}}{\dot{\sigma}}, \quad \ddot{\mathbf{Y}}' = \frac{\ddot{\mathbf{Y}}}{\ddot{\sigma}}, \quad \dddot{\mathbf{Y}}' = \frac{\dddot{\mathbf{Y}} - \ddot{\boldsymbol{\mu}}_y\mathbf{1}^T}{\dddot{\sigma}}$$

are the centered point set matrices. By considering the fact that trace is invariant under cyclic matrix permutations and $\mathbf{r}$ is orthogonal. (12) can be rewritten as:

$$\begin{aligned}Q &= -c_1\mathrm{tr}[(\dot{\mathbf{X}}'\mathbf{P}^T\dot{\mathbf{Y}}'^T)^T\mathbf{r} + (\ddot{\mathbf{X}}'\mathbf{P}^T\ddot{\mathbf{Y}}'^T)^T\mathbf{r} + (\dddot{\mathbf{X}}'\mathbf{P}^T\dddot{\mathbf{Y}}'^T)^T\mathbf{r}] + c_2\\
&= -c_1\mathrm{tr}[(\dot{\mathbf{X}}'\mathbf{P}^T\dot{\mathbf{Y}}'^T + \ddot{\mathbf{X}}'\mathbf{P}^T\ddot{\mathbf{Y}}'^T + \dddot{\mathbf{X}}'\mathbf{P}^T\dddot{\mathbf{Y}}'^T)^T\mathbf{r}] + c_2\end{aligned}$$

where $c_1$, $c_2$ are constants independent of $\mathbf{r}$ and $c_1 > 0$. Thus minimization of $Q$ with respect to $\mathbf{r}$ is equivalent to maximization of

$$\max \mathrm{tr}(\mathbf{A}^T\mathbf{r}), \text{ where}$$
$$A = \dot{\mathbf{X}}'\mathbf{P}^T\hat{\mathbf{Y}}'^T + \ddot{\mathbf{X}}'\mathbf{P}^T\ddot{\mathbf{Y}}'^T + \dddot{\mathbf{X}}'\mathbf{P}^T\dddot{\mathbf{Y}}'^T, \quad \mathbf{r}^T\mathbf{r} = \mathbf{I}, \quad \det(\mathbf{r}) = 1$$

Then, according to *Lemma 1* the optimal $\mathbf{R}$ is in the form

$$\mathbf{r} = \mathbf{U}\mathbf{C}\mathbf{V}^T, \text{ where}$$
$$\mathbf{U}\mathbf{S}\mathbf{S}\mathbf{V}^T = svd(\dot{\mathbf{X}}'\mathbf{P}^T\dot{\mathbf{Y}}'^T + \ddot{\mathbf{X}}'\mathbf{P}^T\ddot{\mathbf{Y}}'^T + \dddot{\mathbf{X}}'\mathbf{P}^T\dddot{\mathbf{Y}}'^T),$$
$$\mathbf{C} = d(1,\ldots,1,\det(\mathbf{U}\mathbf{V}^T))$$

where $svd$ denotes the Singular Value Decomposition (SVD).

*Lemma 1*[16]: Supposing $\mathbf{r}_{D\times D}$ is an unknown rotation matrix, $\mathbf{A}_{D\times D}$ is a known real square matrix, and $\mathbf{USSV}^T$ is a SVD of $\mathbf{A}$. Then the optimal rotation matrix $\mathbf{r}$ that maximizes $\mathrm{tr}(\mathbf{A}^T\mathbf{r})$ is $\mathbf{r} = \mathbf{UCV}^T$, where $\mathbf{C} = d(1,\ldots,1,\det(\mathbf{UV}^T))$.

Other parameters can be updated by equating the partial derivative of to zero, which are all given as a summary in Fig. 1.

---

**Rigid feature matching**

● Initialization:
$\mathbf{r} = \mathbf{I}$, $\mathbf{t} = \mathbf{0}$, $s = 1$, $\omega = 0.1$, $\dot{\sigma}^2 = \frac{1}{2NM}\sum_{n,m=1}^{N,M}\|\dot{\mathbf{x}}_n - \dot{\mathbf{y}}_m\|^2$,

$\ddot{\sigma}^2 = \frac{1}{2NM}\sum_{n,m=1}^{N,M}\|\ddot{\mathbf{x}}_n - \ddot{\mathbf{y}}_m\|^2$, $\dddot{\sigma}^2 = \frac{1}{2NM}\sum_{n,m=1}^{N,M}\|\dddot{\mathbf{x}}_n - \dddot{\mathbf{y}}_m\|^2$.

● Optimization using EM (repeat until convergence):

**E-step**: Compute $\mathbf{P}$:

$$p_{mn} = \frac{\exp(-\frac{1}{2}(\mathbf{x}_n - \mathcal{T}(\mathbf{y}_m,\boldsymbol{\theta}))^T\boldsymbol{\Sigma}^{-1}(\mathbf{x}_n - \mathcal{T}(\mathbf{y}_m,\boldsymbol{\theta})))}{\sum_{m=1}^M \exp(-\frac{1}{2}(\mathbf{x}_n - \mathcal{T}(\mathbf{y}_m,\boldsymbol{\theta}))^T\boldsymbol{\Sigma}^{-1}(\mathbf{x}_n - \mathcal{T}(\mathbf{y}_m,\boldsymbol{\theta}))) + \sqrt{(2\pi)^D|\boldsymbol{\Sigma}|}\frac{\omega}{1-\omega}\frac{M}{N}}$$

**M-step**: Solve for Parameters:

1. $N_p = \mathbf{1}^T\mathbf{P}\mathbf{1}$, $\dot{\mathbf{X}}' = \frac{\dot{\mathbf{X}}}{\dot{\sigma}}$, $\dot{\mathbf{Y}}' = \frac{\dot{\mathbf{Y}}}{\dot{\sigma}}$, $\ddot{\mathbf{X}}' = \frac{\ddot{\mathbf{X}}}{\ddot{\sigma}}$, $\ddot{\mathbf{Y}}' = \frac{\ddot{\mathbf{Y}}}{\ddot{\sigma}}$, $\dddot{\mathbf{X}}' = \frac{\dddot{\mathbf{X}} - \ddot{\boldsymbol{\mu}}'_x\mathbf{1}^T}{\dddot{\sigma}}$,

$\ddot{\boldsymbol{\mu}}_x = \frac{\dddot{\mathbf{X}}\mathbf{P}^T\mathbf{1}}{N_P}$, $\ddot{\boldsymbol{\mu}}_y = \frac{\dddot{\mathbf{Y}}\mathbf{P}\mathbf{1}}{N_P}$, $\dddot{\mathbf{Y}}' = \frac{\dddot{\mathbf{Y}} - \ddot{\boldsymbol{\mu}}_y\mathbf{1}^T}{\dddot{\sigma}}$, $\omega = \frac{(N - N_p)}{N}$;

2. $A = \dot{\mathbf{X}}'\mathbf{P}^T\dot{\mathbf{Y}}'^T + \ddot{\mathbf{X}}'\mathbf{P}^T\ddot{\mathbf{Y}}'^T + \dddot{\mathbf{X}}'\mathbf{P}^T\dddot{\mathbf{Y}}'^T$, $\mathbf{USSV}^T = svd(\mathbf{A})$;

3. $\mathbf{r} = \mathbf{UCV}^T$, where $\mathbf{C} = d(1,\ldots,1,\det(\mathbf{UV}^T))$;

4. $s = \frac{\mathrm{tr}(\dot{\mathbf{X}}'\mathbf{P}^T\dot{\mathbf{Y}}'^T\mathbf{r}^T + \ddot{\mathbf{X}}'\mathbf{P}^T\ddot{\mathbf{Y}}'^T\mathbf{r}^T + \dddot{\mathbf{X}}'\mathbf{P}^T\dddot{\mathbf{Y}}'^T\mathbf{r}^T)}{\mathrm{tr}(\dot{\mathbf{Y}}'d(\mathbf{P}\mathbf{1})\dot{\mathbf{Y}}'^T + \ddot{\mathbf{Y}}'d(\mathbf{P}\mathbf{1})\ddot{\mathbf{Y}}'^T + \dddot{\mathbf{Y}}'d(\mathbf{P}\mathbf{1})\dddot{\mathbf{Y}}'^T)}$;

5. $\mathbf{t} = \ddot{\boldsymbol{\mu}}_x - s\mathbf{r}\ddot{\boldsymbol{\mu}}_y$;

6. $\dot{\sigma}^2 = \frac{1}{2N_p}\mathrm{tr}[\dot{\mathbf{X}}d(\mathbf{P}^T\mathbf{1})\dot{\mathbf{X}}^T - 2s\dot{\mathbf{X}}\mathbf{P}^T\dot{\mathbf{Y}}^T\mathbf{r}^T + s^2\dot{\mathbf{Y}}d(\mathbf{P}\mathbf{1})\dot{\mathbf{Y}}^T]$,

$\ddot{\sigma}^2 = \frac{1}{2N_p}\mathrm{tr}[\ddot{\mathbf{X}}d(\mathbf{P}^T\mathbf{1})\ddot{\mathbf{X}}^T - 2s\ddot{\mathbf{X}}\mathbf{P}^T\ddot{\mathbf{Y}}^T\mathbf{r}^T + s^2\ddot{\mathbf{Y}}d(\mathbf{P}\mathbf{1})\ddot{\mathbf{Y}}^T]$,

$\dddot{\sigma}^2 = \frac{1}{2N_p}\mathrm{tr}[\dddot{\mathbf{X}}'d(\mathbf{P}^T\mathbf{1})\dddot{\mathbf{X}}'^T - 2s\dddot{\mathbf{X}}'\mathbf{P}^T\dddot{\mathbf{Y}}'^T\mathbf{r}^T + s^2\dddot{\mathbf{Y}}'d(\mathbf{P}\mathbf{1})\dddot{\mathbf{Y}}'^T]$.

The matched point set: $\mathcal{T}(\mathbf{Y}) = sr\dot{\mathbf{Y}} + \mathbf{t}$;
Correspondence probability is given by $\mathbf{P}$

**Fig. 1** Rigid feature matching.

### B. Affine Feature Matching

Compared with the rigid case, affine feature matching is simpler since the optimization is unconstrained. The affine transformation is defined as $\mathcal{T}(\mathbf{y}_m) = \mathbf{B}\mathbf{y}_m + \mathbf{t}$, where $\mathbf{B}$ is a $2\times 2$ affine matrix indicating of feature shape, scale and orientation, and $\mathbf{t}$ is the translation vector. The objective function (10) then becomes

$$\begin{aligned}Q(\mathbf{B},\mathbf{t},\boldsymbol{\Sigma}) = \frac{1}{2}\sum_{n,m=1}^{N,M} p_{mn}[&(\mathbf{x}_n - \mathbf{B}\mathbf{y}_m - \mathbf{t})^T\boldsymbol{\Sigma}^{-1}(\mathbf{x}_n - \mathbf{B}\mathbf{y}_m - \mathbf{t})]\\
&+ N_p\log|\boldsymbol{\Sigma}| + \frac{N_pD}{2}\log(2\pi) - N_p\log(\frac{1-\omega}{M})\\
&- (N - N_p)\log(\frac{\omega}{N})\end{aligned} \quad (13)$$

Besides, we also consider:

$$(\mathbf{x}_n - \mathbf{B}\mathbf{y}_m - \mathbf{t})^T\boldsymbol{\Sigma}^{-1}(\mathbf{x}_n - \mathbf{B}\mathbf{y}_m - \mathbf{t}) =$$
$$\left\|\frac{\dot{\mathbf{x}}_n - \mathbf{B}\dot{\mathbf{y}}_m}{\dot{\sigma}}\right\|^2 + \left\|\frac{\ddot{\mathbf{x}}_n - \mathbf{B}\ddot{\mathbf{y}}_m}{\ddot{\sigma}}\right\|^2 + \left\|\frac{\dddot{\mathbf{x}}_n - \mathbf{B}\dddot{\mathbf{y}}_m - \mathbf{t}^{2\times 1}}{\dddot{\sigma}}\right\|^2 \quad (14)$$

Substituting (14) to the $Q$ function, it becomes:

$$Q(\mathbf{B},\mathbf{t},\boldsymbol{\Sigma}) = \frac{N_p D}{2}\log(2\pi) + 2N_p \log(\dot{\sigma}\ddot{\sigma}\dddot{\sigma})$$
$$+ \sum_{n,m=1}^{N,M} \frac{p_{mn}}{2}(\left\|\frac{\dot{\mathbf{x}}_n - \mathbf{B}\dot{\mathbf{y}}_m}{\dot{\sigma}}\right\|^2 + \left\|\frac{\ddot{\mathbf{x}}_n - \mathbf{B}\ddot{\mathbf{y}}_m}{\ddot{\sigma}}\right\|^2 + \left\|\frac{\dddot{\mathbf{x}}_n - \mathbf{B}\dddot{\mathbf{y}}_m - \mathbf{t}^{2\times 1}}{\dddot{\sigma}}\right\|^2) \quad (15)$$
$$- N_p \log(1-\omega) - (N - N_p)\log(\omega)$$

The solution of $\mathbf{t}$ is similar to the rigid case. The solution of $\mathbf{B}$ can be obtained by directly taking the partial derivative of $Q$, setting it to zero, and solving the resulting linear system of equations. The solutions are given as follows and we also summarize the algorithm in Fig. 3.

$$\mathbf{t} = \frac{1}{N_P}\dddot{\mathbf{X}}\mathbf{P}^T\mathbf{1} - \mathbf{B}\frac{1}{N_P}\dddot{\mathbf{Y}}\mathbf{P1} = \dddot{\boldsymbol{\mu}}_x - \mathbf{B}\dddot{\boldsymbol{\mu}}_y$$

Substituting $\mathbf{t}$ back into the objective function and rewriting it in matrix form, we obtain

$$Q = \frac{1}{2}\text{tr}[\dot{\mathbf{X}}'d(\mathbf{P}^T\mathbf{1})\dot{\mathbf{X}}'^T - 2\dot{\mathbf{X}}'\mathbf{P}^T\dot{\mathbf{Y}}'^T\mathbf{B}^T + \mathbf{B}\dot{\mathbf{Y}}'d(\mathbf{P1})\dot{\mathbf{Y}}'^T\mathbf{B}^T$$
$$+ \ddot{\mathbf{X}}'d(\mathbf{P}^T\mathbf{1})\ddot{\mathbf{X}}'^T - 2\ddot{\mathbf{X}}'\mathbf{P}^T\ddot{\mathbf{Y}}'^T\mathbf{B}^T + \mathbf{B}\ddot{\mathbf{Y}}'d(\mathbf{P1})\ddot{\mathbf{Y}}'^T\mathbf{B}^T \quad (16)$$
$$+ \dddot{\mathbf{X}}'d(\mathbf{P}^T\mathbf{1})\dddot{\mathbf{X}}'^T - 2\dddot{\mathbf{X}}'\mathbf{P}^T\dddot{\mathbf{Y}}'^T\mathbf{B}^T + \mathbf{B}\dddot{\mathbf{Y}}'d(\mathbf{P1})\dddot{\mathbf{Y}}'^T\mathbf{B}^T]$$
$$+ 2N_p\log(\dot{\sigma}\ddot{\sigma}\dddot{\sigma})$$

where $\dot{\mathbf{X}}' = \dot{\mathbf{X}}/\dot{\sigma}$, $\dot{\mathbf{Y}}' = \dot{\mathbf{Y}}/\dot{\sigma}$, $\ddot{\mathbf{X}}' = \ddot{\mathbf{X}}/\ddot{\sigma}$, $\ddot{\mathbf{Y}}' = \ddot{\mathbf{Y}}/\ddot{\sigma}$, $\dddot{\mathbf{X}}' = (\dddot{\mathbf{X}} - \dddot{\boldsymbol{\mu}}_x'\mathbf{1}^T)/\dddot{\sigma}$ and $\dddot{\mathbf{Y}}' = (\dddot{\mathbf{Y}} - \dddot{\boldsymbol{\mu}}_y'\mathbf{1}^T)/\dddot{\sigma}$ are the centered point set matrices. Then the partial derivatives of $Q$ with respect to $\mathbf{B}$ is:

$$\mathbf{B} = (\dot{\mathbf{X}}'\mathbf{P}^T\dot{\mathbf{Y}}'^T + \ddot{\mathbf{X}}'\mathbf{P}^T\ddot{\mathbf{Y}}'^T + \dddot{\mathbf{X}}'\mathbf{P}^T\dddot{\mathbf{Y}}'^T) \cdot$$
$$(\dot{\mathbf{Y}}'d(\mathbf{P1})\dot{\mathbf{Y}}'^T + \ddot{\mathbf{Y}}'d(\mathbf{P1})\ddot{\mathbf{Y}}'^T + \dddot{\mathbf{Y}}'d(\mathbf{P1})\dddot{\mathbf{Y}}'^T)^{-1}$$

And the whole algorithm is given in Fig. 2.

---

**Affine feature matching**

- **Initialization:**

  $\mathbf{B} = \mathbf{I}$, $\mathbf{t} = \mathbf{0}$, $\omega = 0.1$, $\dot{\sigma}^2 = \frac{1}{2NM}\sum_{n,m=1}^{N,M}\|\dot{\mathbf{x}}_n - \dot{\mathbf{y}}_m\|^2$,

  $\ddot{\sigma}^2 = \frac{1}{2NM}\sum_{n,m=1}^{N,M}\|\ddot{\mathbf{x}}_n - \ddot{\mathbf{y}}_m\|^2$, $\dddot{\sigma}^2 = \frac{1}{2NM}\sum_{n,m=1}^{N,M}\|\dddot{\mathbf{x}}_n - \dddot{\mathbf{y}}_m\|^2$.

- **Optimization using EM (repeat until convergence):**
  **E-step:** Compute $\mathbf{P}$:

  $$p_{mn} = \frac{\exp(-\frac{1}{2}(\mathbf{x}_n - \mathcal{T}(\mathbf{y}_m,\boldsymbol{\theta}))^T \boldsymbol{\Sigma}^{-1}(\mathbf{x}_n - \mathcal{T}(\mathbf{y}_m,\boldsymbol{\theta})))}{\sum_{m=1}^M \exp(-\frac{1}{2}(\mathbf{x}_n - \mathcal{T}(\mathbf{y}_m,\boldsymbol{\theta}))^T \boldsymbol{\Sigma}^{-1}(\mathbf{x}_n - \mathcal{T}(\mathbf{y}_m,\boldsymbol{\theta}))) + \sqrt{(2\pi)^D}|\boldsymbol{\Sigma}|\frac{\omega}{1-\omega}\frac{M}{N}}$$

  **M-step:** Solve for Parameters:

  1. $N_p = \mathbf{1}^T\mathbf{P1}$, $\dot{\mathbf{X}}' = \frac{\dot{\mathbf{X}}}{\dot{\sigma}}$, $\dot{\mathbf{Y}}' = \frac{\dot{\mathbf{Y}}}{\dot{\sigma}}$, $\ddot{\mathbf{X}}' = \frac{\ddot{\mathbf{X}}}{\ddot{\sigma}}$, $\ddot{\mathbf{Y}}' = \frac{\ddot{\mathbf{Y}}}{\ddot{\sigma}}$, $\dddot{\mathbf{X}}' = \frac{\dddot{\mathbf{X}} - \dddot{\boldsymbol{\mu}}_x\mathbf{1}^T}{\dddot{\sigma}}$,

     $\dddot{\boldsymbol{\mu}}_x = \frac{\dddot{\mathbf{X}}\mathbf{P}^T\mathbf{1}}{N_p}$, $\dddot{\boldsymbol{\mu}}_y = \frac{\dddot{\mathbf{Y}}\mathbf{P1}}{N_p}$, $\dddot{\mathbf{Y}}' = \frac{\dddot{\mathbf{Y}} - \dddot{\boldsymbol{\mu}}_y\mathbf{1}^T}{\dddot{\sigma}}$, $\omega = \frac{(N - N_p)}{N}$;

  2. $\mathbf{B} = (\dot{\mathbf{X}}'\mathbf{P}^T\dot{\mathbf{Y}}'^T + \ddot{\mathbf{X}}'\mathbf{P}^T\ddot{\mathbf{Y}}'^T + \dddot{\mathbf{X}}'\mathbf{P}^T\dddot{\mathbf{Y}}'^T) \cdot$
     $(\dot{\mathbf{Y}}'d(\mathbf{P1})\dot{\mathbf{Y}}'^T + \ddot{\mathbf{Y}}'d(\mathbf{P1})\ddot{\mathbf{Y}}'^T + \dddot{\mathbf{Y}}'d(\mathbf{P1})\dddot{\mathbf{Y}}'^T)^{-1}$

  3. $\mathbf{t} = \dddot{\boldsymbol{\mu}}_x - \mathbf{B}\dddot{\boldsymbol{\mu}}_y$;

  4. $\dot{\sigma}^2 = \frac{1}{2N_p}\text{tr}[\dot{\mathbf{X}}d(\mathbf{P}^T\mathbf{1})\dot{\mathbf{X}}^T - 2\dot{\mathbf{X}}\mathbf{P}^T\dot{\mathbf{Y}}^T\mathbf{B}^T + \mathbf{B}\dot{\mathbf{Y}}d(\mathbf{P1})\dot{\mathbf{Y}}^T\mathbf{B}^T]$,

     $\ddot{\sigma}^2 = \frac{1}{2N_p}\text{tr}[\ddot{\mathbf{X}}d(\mathbf{P}^T\mathbf{1})\ddot{\mathbf{X}}^T - 2\ddot{\mathbf{X}}\mathbf{P}^T\ddot{\mathbf{Y}}^T\mathbf{B}^T + \mathbf{B}\ddot{\mathbf{Y}}d(\mathbf{P1})\ddot{\mathbf{Y}}^T\mathbf{B}^T]$,

     $\dddot{\sigma}^2 = \frac{1}{2N_p}\text{tr}[\dddot{\mathbf{X}}'d(\mathbf{P}^T\mathbf{1})\dddot{\mathbf{X}}'^T - 2\dddot{\mathbf{X}}'\mathbf{P}^T\dddot{\mathbf{Y}}'^T\mathbf{B}^T + \mathbf{B}\dddot{\mathbf{Y}}'d(\mathbf{P1})\dddot{\mathbf{Y}}'^T\mathbf{B}^T]$.

The matched point set: $\mathcal{T}(\dot{\mathbf{Y}}) = \mathbf{B}\dot{\mathbf{Y}} + \mathbf{t}$;
Correspondence probability is given by $\mathbf{P}$

**Fig. 2** Affine feature matching.

*C. Non-rigid Feature Matching*

In the case of non-rigid matching, we define the transformation $\mathcal{T}$ as the original position plus a displacement function [1, 17]. Suppose $\mathcal{V}(\mathbf{Y})$ is the displacement function of the points in $\mathbf{Y}$ then $\mathcal{T}(\mathbf{Y},\mathcal{V}) = \mathbf{Y} + \mathcal{V}(\mathbf{Y})$ is the new location of $\mathbf{Y}$. Here, we define the prior over $\mathcal{V}$ in the Tikhonov regularization framework as [16]:

$$p(v) = \exp(-\frac{\lambda}{2}\|\mathcal{V}\|_{\mathcal{H}}^2) \quad (17)$$

where $\|\mathcal{V}\|_{\mathcal{H}}^2$ is the norm of $\mathcal{V}(\mathbf{Y})$ in the Reproduction Kernel Hilbert Space (RKHS). Intuitively, the smaller $\|\mathcal{V}\|_{\mathcal{H}}^2$ is, the more coherent the motion between two point sets will be. The optimal form of $\mathcal{V}$ can be written as the linear combination of kernels:

$$\mathcal{V}(\mathbf{z}) = \sum_{m=1}^M w_m G(\mathbf{z},\mathbf{y}_m) = \mathbf{WG} = \begin{vmatrix} \dot{\mathbf{W}}\dot{\mathbf{G}} \\ \ddot{\mathbf{W}}\ddot{\mathbf{G}} \\ \dddot{\mathbf{W}}\dddot{\mathbf{G}} \end{vmatrix} \quad (18)$$

where $\mathbf{W}^{D \times M} = (\mathbf{w}_1, ..., \mathbf{w}_M)$ is a matrix of coefficients, $\mathbf{G}$ is a $m \times m$ kernel matrix with $G(\mathbf{y}_i, \mathbf{y}_j) = \exp(-\frac{1}{2}\frac{\|\mathbf{y}_i - \mathbf{y}_j\|^2}{\beta})$. By multiplying the likelihood (2) by the priori (5), we can get the posteriori

$$p(\theta, \omega | \mathbf{Y}) \propto L(\theta, \sigma^2) p(\mathcal{V}) \quad (19)$$

The parameters could be estimated by solving the minimizing the negative log-posterior:

$$Q(\mathcal{V}, \Sigma) = \frac{N_p D}{2} \log(2\pi) - (N - N_p) \log(\frac{\omega}{N}) - N_p \log(\frac{1-\omega}{M})$$
$$+ \sum_{n,m=1}^{N,M} \frac{p_{mn}}{2} (\mathbf{x}_n - (\mathbf{y}_m + \mathcal{V}(\mathbf{y}_m)))^T \Sigma^{-1} (\mathbf{x}_n - (\mathbf{y}_m + \mathcal{V}(\mathbf{y}_m))) \quad (20)$$
$$+ N_p \log(|\Sigma|) + \frac{\lambda}{2} tr(\mathbf{W G W}^T)$$

And let $\dot{\mathbf{T}} = \mathcal{T}(\dot{\mathbf{Y}}, \mathcal{V}) = \dot{\mathbf{Y}} + \dot{\mathbf{W}}\dot{\mathbf{G}}$, $\ddot{\mathbf{T}} = \mathcal{T}(\ddot{\mathbf{Y}}, \mathcal{V}) = \ddot{\mathbf{Y}} + \ddot{\mathbf{W}}\ddot{\mathbf{G}}$ and $\dddot{\mathbf{T}} = \mathcal{T}(\dddot{\mathbf{Y}}, \mathcal{V}) = \dddot{\mathbf{Y}} + \dddot{\mathbf{W}}\dddot{\mathbf{G}}$; we can also write (20) as:

$$Q(\mathcal{V}, \Sigma) = \frac{N_p D}{2} \log(2\pi) - N_p \log(\frac{1-\omega}{M}) - (N - N_p) \log(\frac{\omega}{N})$$
$$+ \sum_{n,m=1}^{N,M} \frac{p_{mn}}{2} [\left\|\frac{\dot{\mathbf{x}}_n - \dot{\mathbf{T}}_m}{\dot{\sigma}}\right\|^2 + \left\|\frac{\ddot{\mathbf{x}}_n - \ddot{\mathbf{T}}_m}{\ddot{\sigma}}\right\|^2 + \left\|\frac{\dddot{\mathbf{x}}_n - \dddot{\mathbf{T}}_m}{\dddot{\sigma}}\right\|^2] \quad (21)$$
$$+ 2N_p \log(\dot{\sigma}\ddot{\sigma}\dddot{\sigma}) + \frac{\lambda}{2} tr(\dot{\mathbf{W}}\dot{\mathbf{G}}\dot{\mathbf{W}}^T + \ddot{\mathbf{W}}\ddot{\mathbf{G}}\ddot{\mathbf{W}}^T + \dddot{\mathbf{W}}\dddot{\mathbf{G}}\dddot{\mathbf{W}}^T)$$

Then, let $\frac{\partial Q}{\partial \mathbf{W}} = 0$, we can solve $\mathbf{W}$ from:

$$\dot{\mathbf{W}}(\dot{\mathbf{G}} d(\mathbf{P1}) + \lambda \dot{\sigma}^2) = \dot{\mathbf{X}} \mathbf{P} - \dot{\mathbf{Y}} d(\mathbf{P1})$$
$$\ddot{\mathbf{W}}(\ddot{\mathbf{G}} d(\mathbf{P1}) + \lambda \ddot{\sigma}^2) = \ddot{\mathbf{X}} \mathbf{P} - \ddot{\mathbf{Y}} d(\mathbf{P1})$$
$$\dddot{\mathbf{W}}(\dddot{\mathbf{G}} d(\mathbf{P1}) + \lambda \dddot{\sigma}^2) = \dddot{\mathbf{X}} \mathbf{P} - \dddot{\mathbf{Y}} d(\mathbf{P1})$$

Other parameters can also obtained by equating the corresponding derivative of $Q$ to zeros.

---

**Non-rigid feature matching**

● Initialization:
$\mathbf{W} = 0$, $\beta = 2$, $\lambda = 3$, $\omega = 0.1$, $\dot{\sigma}^2 = \frac{1}{2NM}\sum_{n,m=1}^{N,M}\|\dot{\mathbf{x}}_n - \dot{\mathbf{y}}_m\|^2$,
$\ddot{\sigma}^2 = \frac{1}{2NM}\sum_{n,m=1}^{N,M}\|\ddot{\mathbf{x}}_n - \ddot{\mathbf{y}}_m\|^2$, $\dddot{\sigma}^2 = \frac{1}{2NM}\sum_{n,m=1}^{N,M}\|\dddot{\mathbf{x}}_n - \dddot{\mathbf{y}}_m\|^2$.

● Optimization using EM (repeat until convergence):
**E-step**: Compute $\mathbf{P}$:
$$p_{mn} = \frac{\exp(-\frac{1}{2}(\mathbf{x}_n - \mathcal{T}(\mathbf{y}_m, \theta))^T \Sigma^{-1}(\mathbf{x}_n - \mathcal{T}(\mathbf{y}_m, \theta)))}{\sum_{m=1}^{M}\exp(-\frac{1}{2}(\mathbf{x}_n - \mathcal{T}(\mathbf{y}_m, \theta))^T \Sigma^{-1}(\mathbf{x}_n - \mathcal{T}(\mathbf{y}_m, \theta))) + \sqrt{(2\pi)^D}|\Sigma|\frac{\omega}{1-\omega}\frac{M}{N}}$$

**M-step**: Solve for Parameters:
1. $\dot{\mathbf{T}} = \dot{\mathbf{Y}} + \dot{\mathbf{W}}\dot{\mathbf{G}}$, $\ddot{\mathbf{T}} = \ddot{\mathbf{Y}} + \ddot{\mathbf{W}}\ddot{\mathbf{G}}$, $\dddot{\mathbf{T}} = \dddot{\mathbf{Y}} + \dddot{\mathbf{W}}\dddot{\mathbf{G}}$,
2. $N_p = \mathbf{1}^T \mathbf{P1}$, $\omega = \frac{(N - N_p)}{N}$;
3. solve $\dot{\mathbf{W}}$, $\ddot{\mathbf{W}}$ and $\dddot{\mathbf{W}}$ according to:
   $\dot{\mathbf{W}}(\dot{\mathbf{G}}diag(\mathbf{P1}) + \lambda\dot{\sigma}^2) = \dot{\mathbf{X}}\mathbf{P} - \dot{\mathbf{Y}}diag(\mathbf{P1})$,
   $\ddot{\mathbf{W}}(\ddot{\mathbf{G}}diag(\mathbf{P1}) + \lambda\ddot{\sigma}^2) = \ddot{\mathbf{X}}\mathbf{P} - \ddot{\mathbf{Y}}diag(\mathbf{P1})$,
   $\dddot{\mathbf{W}}(\dddot{\mathbf{G}}diag(\mathbf{P1}) + \lambda\dddot{\sigma}^2) = \dddot{\mathbf{X}}\mathbf{P} - \dddot{\mathbf{Y}}diag(\mathbf{P1})$;
4. $\dot{\sigma}^2 = \frac{1}{N_p D}tr(\dot{\mathbf{X}}d(\mathbf{P}^T\mathbf{1})\dot{\mathbf{X}}^T - 2\dot{\mathbf{X}}\mathbf{P}^T\dot{\mathbf{T}}^T + \dot{\mathbf{T}}d(\mathbf{P1})\dot{\mathbf{T}}^T)$
   $\ddot{\sigma}^2 = \frac{1}{N_p D}tr(\ddot{\mathbf{X}}d(\mathbf{P}^T\mathbf{1})\ddot{\mathbf{X}}^T - 2\ddot{\mathbf{X}}\mathbf{P}^T\ddot{\mathbf{T}}^T + \ddot{\mathbf{T}}d(\mathbf{P1})\ddot{\mathbf{T}}^T)$
   $\dddot{\sigma}^2 = \frac{1}{N_p D}tr(\dddot{\mathbf{X}}d(\mathbf{P}^T\mathbf{1})\dddot{\mathbf{X}}^T - 2\dddot{\mathbf{X}}\mathbf{P}^T\dddot{\mathbf{T}}^T + \dddot{\mathbf{Y}}d(\mathbf{P1})\dddot{\mathbf{T}}^T)$

The matched point set: $\mathcal{T}(\dot{\mathbf{Y}}, \mathcal{V}) = \dot{\mathbf{Y}} + \dot{\mathbf{W}}\dot{\mathbf{G}}$
Correspondence probability is given by $\mathbf{P}$

**Fig. 3** Non-rigid feature matching.

*D. Implementation Details*

Parameter Setting: There are three parameters in the proposed method: $\omega$, $\lambda$ and $\beta$, which are all the necessary parameters in CPD-based framework. $\omega$ is the initial assumption of the amount of outliers in the point sets. Parameter $\beta$ and $\lambda$ are used in non-rigid version, where the former represents the range of the interaction between feature points, and the latter represents the trade-off between the goodness of maximum likelihood fit and regularization. We set $\omega = 0.1$, $\lambda = 3$ and $\beta = 2$ throughout the paper. In addition, a threshold 0.8 is used as the possibility threshold to judge the correctness of a correspondence, and we stop the algorithms when the difference of the objective function value is less than $10^{-5}$.

IV. EXPERIMENTS

*A. Experimental Setup*

*1) Dataset*

To evaluate our model comprehensively, we establish a datasets of 14 image pairs composed of three sub-sets: rigid set (pair 1 to 5), affine set (pair 6 to 10), and non-rigid set (pair 11 to 14). And both optical and SAR images are involved. In the rigid and affine set, the ten image pairs are all chosen from remote sensing dataset as the remote sensing image pairs generally satisfy the global transformation model. It is note that the two images in each pair are captured in different time with significant changes, and the image quality is degraded especially in the SAR images because of speckle noise. These factors result in the increase of matching challenge. The non-rigid set is consists of five image pairs with multi-objects and significant distortions. All the images are shown in Fig. 4, where the ground truth is obtainable.

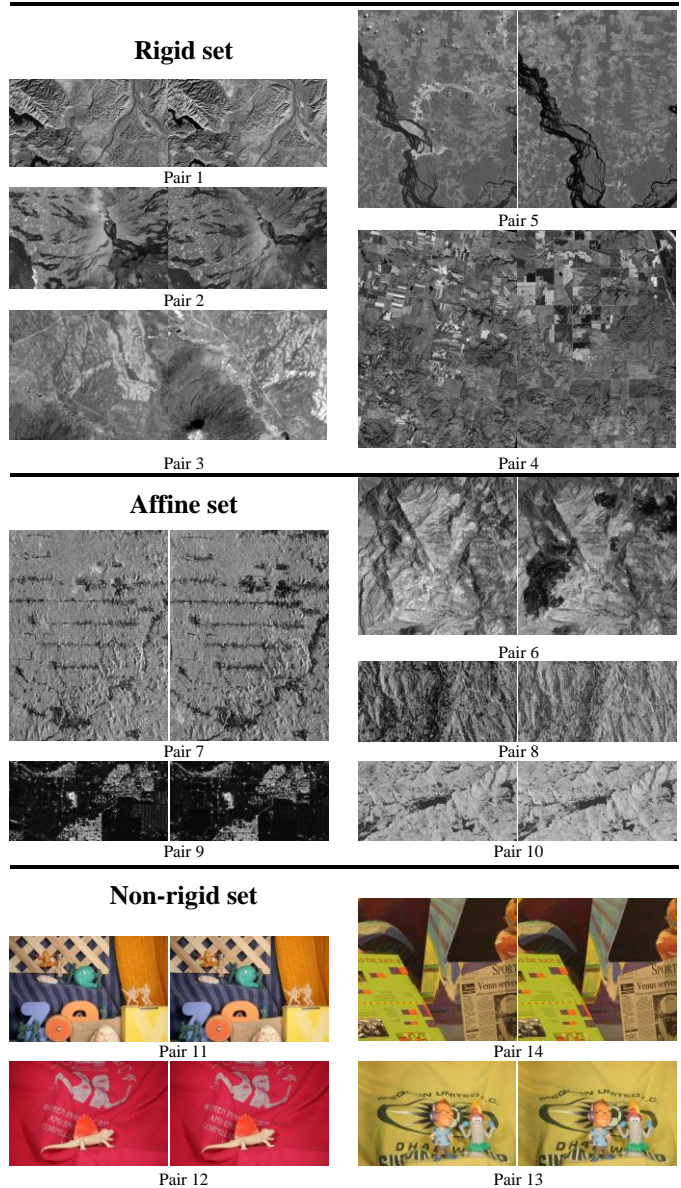

**Fig. 4** Test image pairs.

*2) Compared Algorithms*

Here, we compare the proposed method with CPD [16] and NGMM [17], where CPD is either a well-known GMM-based algorithm using location information only or the basic study of our algorithm, NGMM is an improved version of CPD which combined the information of location and feature descriptor.

*3) Other Experimental Details*

We use the popular DoG [7] detector (based on the VLFeat toolbox [32]). The results are evaluated by precision, recall or F1 ( $F1 = \frac{2 \times precision \times recall}{precision + recall}$ ). The ground truth, i.e., determining the matching correctness of each correspondence, is established according to the rough results of SIFT, with the results confirmed after filtering by the repeatability evaluation criterion given in [6]. The experiments are conducted using MATLAB code with 3.4-GHz Intel Core CPU and 8-GB memory.

*B. Robustness Test*

In this subsection, we test the robustness of the proposed algorithm with respect to outliers. We firstly choose one pair from each subset. For each pair, 200 random correct matches are firstly selected as the ground truth as shown in Fig. 5(a) (blue lines). Then, we add outliers to both point sets selected randomly from false matches between the two images. And we conduct each experiment by gradually increase the number of outliers. 30 trials are conducted with the same settings, and the mean and variance are calculated.

From the figure, we see that, in general, both CPD and the proposed algorithm can obtain high F1 value when there are few outliers, e.g., in the case of 15%. However, the results of CPD drop immediately when the outlier-percentage is over 30% on all three image pairs. This means the failure of CPD in matching. In contrast, the proposed algorithm not only obtains better results in the low-outlier cases on pair 1 and pair 11, but also declines much more slowly than CPD with the outlier ratio increases from 30% to 50%. Besides, the iteration number of the proposed method increases very slowly as the outliers increase. This demonstrates that the efficiency of our method is good balanced with the robustness.

*C. Total Precision-Recalls on the Dataset*

We next give the results on the whole dataset. Three tests are conducted: 1) Rigid test, 2) Affine test, 3) non-rigid test, where each test is conducted on the corresponding sub-dataset using the corresponding model of our method. The results are organized in forms of precision-recall pairs in Fig. 6, where each scattered dot indicates both the average precision and recall of 30 runs on an image pair. From Fig. 6a, we see that all algorithms produce satisfying precision in rigid test; but the proposed method shows better recall than CPD and NGMM. In affine test (Fig. 6b) and non-rigid test (Fig. 6c), we can even find larger gaps between our methods and the compared ones. Specifically, CPD and NGMM never obtain a recall value higher than 0.8 and 0.5 in affine test and non-rigid test respectively, and all the precision values are lower than 0.9 in the tests. In contrast, our proposed method haves much better matching performance, where almost all values are close to 1.

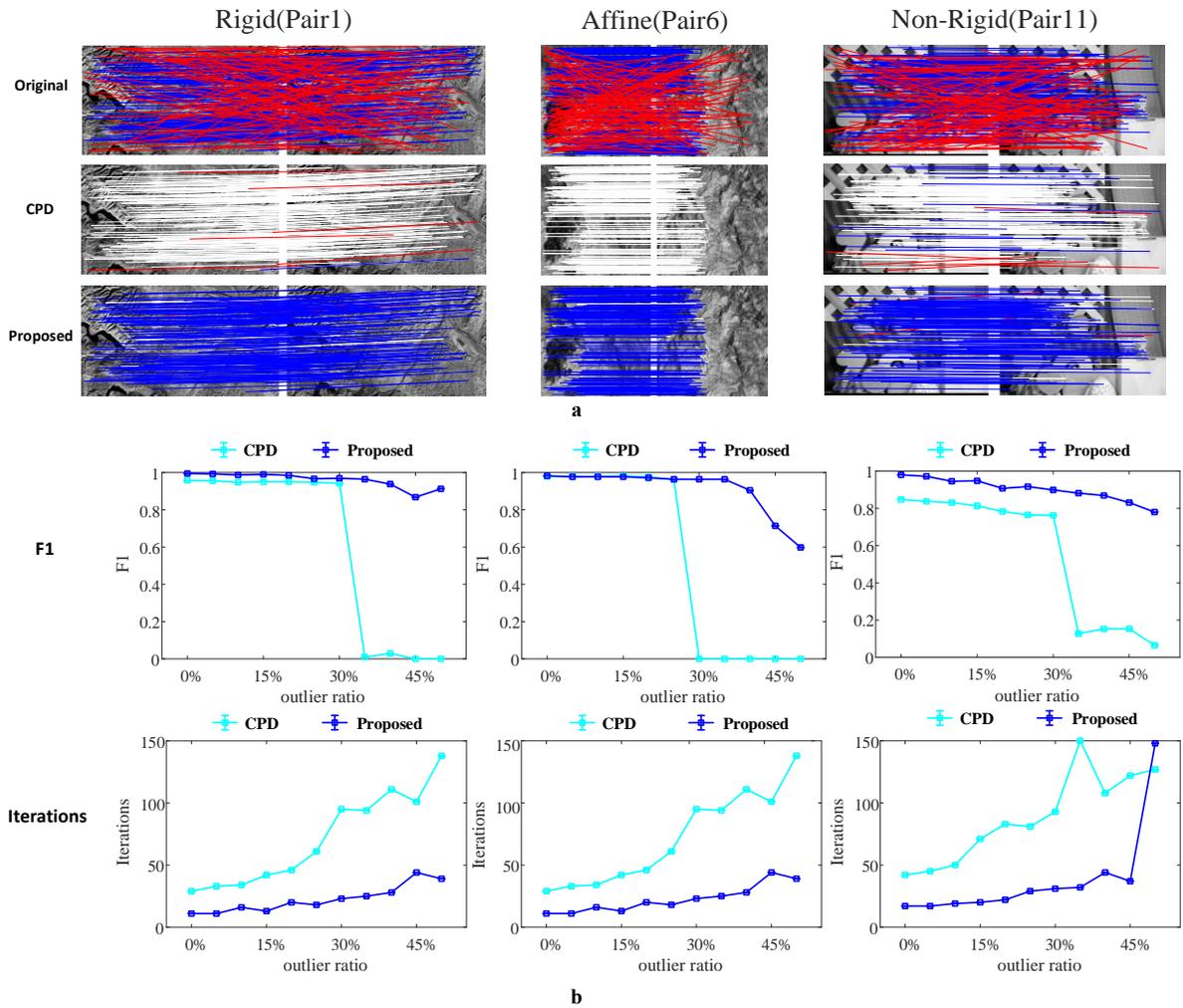

**Fig. 5** Outlier tests: (a) from left to right: image pairs 1, 6 and 11 and some corresponding results in 35% outlier ratio with the marked correct matches (blue), false matches (red) and missing matches (white). (b) From left to right: The F1 results and corresponding iteration number with increasing outlier ratio on image pair 1, 6 and 11 (corresponding to rigid, affine and non-rigid case).

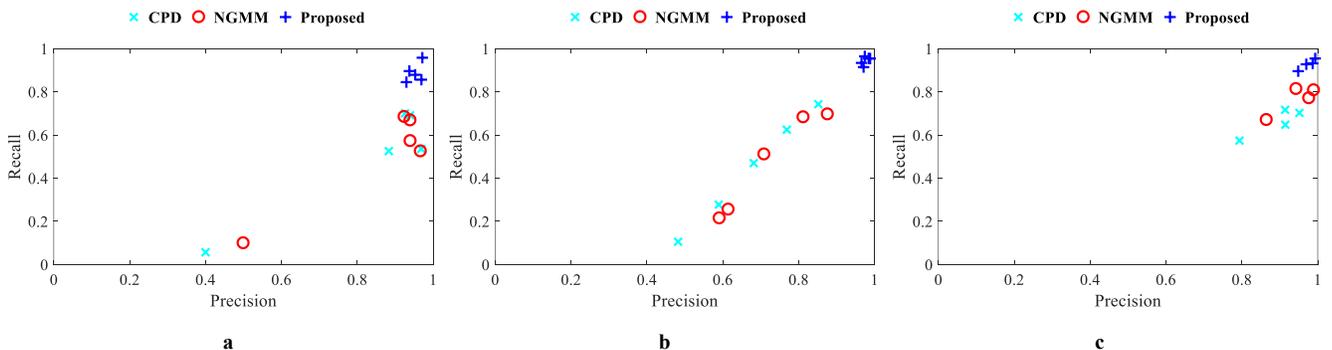

**Fig. 6** Average precision–recall pairs on the datasets. From left to right: a. rigid model on rigid sub-dataset, b. affine model on affine sub-dataset; c. non-rigid model on non-rigid sub-dataset.

### D. Efficiency Analysis

This subsection deals with the efficiency performance. We recorded the average iteration number and run time in Table I (based on the test in subsection C). It can be seen that the proposed algorithm takes much less iteration number and time than CPD and NGMM in all tests especially on image pair 1, 2 and 13, where the proposed algorithm costs no more than one-third of iteration numbers CPD and NGMM takes. This leads to the better efficiency of our algorithm in terms of run time except the case on pair 3 and 4, where CPD and NGMM need same time to converge. We find that the iteration number is very stable in all cases without any variance among 30 runs, and the variance values of run time are also negligible compared with the corresponding mean values.

To further illustrate the convergence details, an iterative process of image pair 2 is shown in Fig. 8. The algorithms start from same original condition and same certain iteration conditions are displayed for comparison. The results show that the proposed algorithm is about to converge in the 11$^{th}$ iteration and nearly converges in 13$^{th}$ iteration. Comparatively, CPD is far from convergence in corresponding iteration numbers, and visually speaking, the result of CPD in 26$^{th}$ iteration is even worse than that of our algorithm in 11$^{th}$ iteration.

TABLE I
AVERAGE ITERATION NUMBER AND RUN TIME ON THE WHOLE DATASET

| Pair | CPD | | NGMM | | Proposed | |
|---|---|---|---|---|---|---|
| | Iter. Num. | Run Time (s) | Iter. Num. | Run Time (s) | Iter. Num. | Run Time (s) |
| 1 | 64 | 1.71±0.014 | 62 | 1.89±0.006 | 18 | 1.09±0.020 |
| 2 | 150[a] | 1.90±0.014 | 138 | 2.06±0.012 | 26 | 0.97±0.013 |
| 3 | 33 | 0.05±0.009 | 35 | 0.08±0.006 | 15 | 0.05±0.009 |
| 4 | 55 | 0.02±0.001 | 52 | 0.03±0.004 | 25 | 0.02±0.001 |
| 5 | 50 | 0.35±0.016 | 49 | 0.51±0.016 | 22 | 0.21±0.015 |
| 6 | 150 | 3.54±0.023 | 108 | 3.41±0.007 | 45 | 2.19±0.058 |
| 7 | 126 | 1.85±0.008 | 105 | 1.96±0.010 | 50 | 1.20±0.017 |
| 8 | 113 | 1.03±0.009 | 100 | 1.14±0.006 | 51 | 0.71±0.013 |
| 9 | 150 | 0.96±0.016 | 150 | 1.16±0.006 | 77 | 0.54±0.028 |
| 10 | 90 | 0.35±0.009 | 67 | 0.37±0.009 | 44 | 0.19±0.011 |
| 11 | 93 | 1.15±0.011 | 104 | 1.39±0.010 | 52 | 1.03±0.015 |
| 12 | 150 | 1.10±0.009 | 150 | 1.29±0.005 | 56 | 0.74±0.034 |
| 13 | 150 | 1.38±0.005 | 121 | 1.46±0.005 | 31 | 0.79±0.036 |
| 14 | 88 | 0.78±0.022 | 76 | 0.83±0.011 | 60 | 0.71±0.028 |

[a]150 is set as the maximum iteration number.

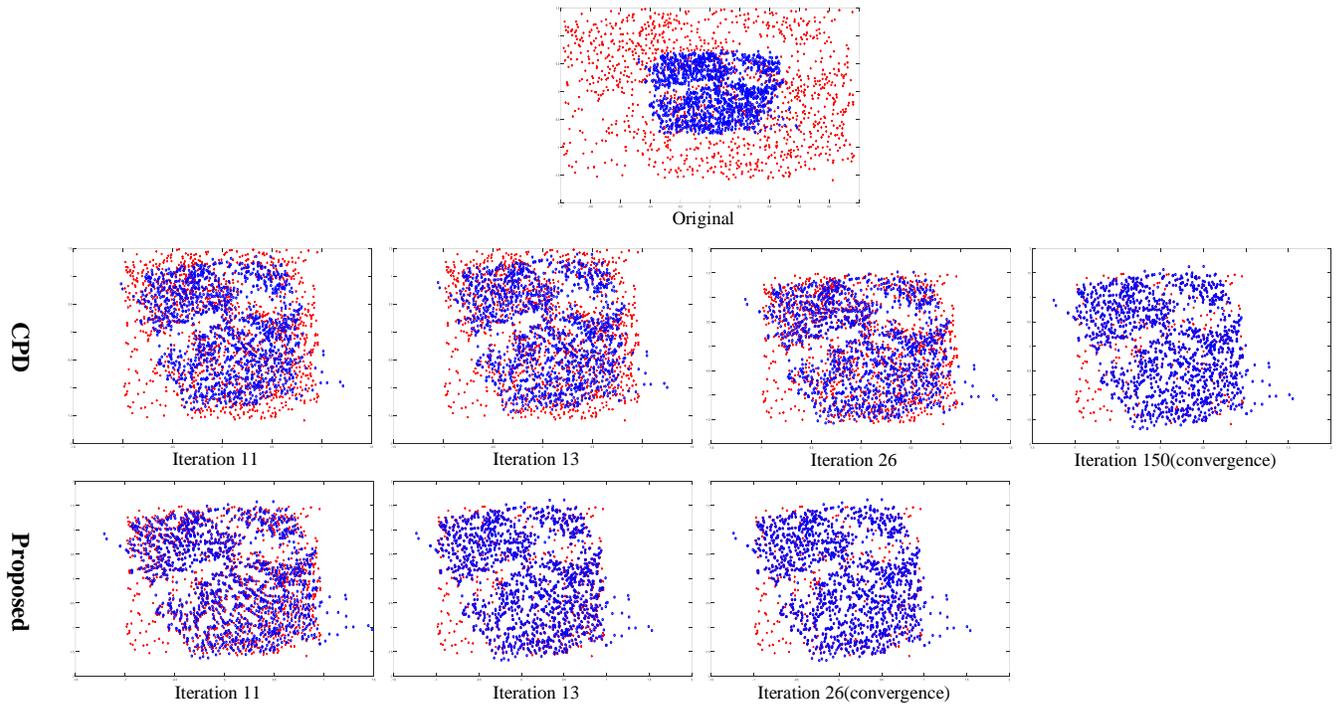

Fig. 8 the iterative processes of CPD and proposed matching method on image pair 2. Top row: the original condition; Second row: the result of CPD. Bottom row: the results by the proposed method.

## V. CONCLUSION

A novel Gaussian Mixture Model (GMM) based point matching algorithm is proposed for co-variant feature matching. Different from the traditional methods which only use the point coordinates the strategy we propose can integrate and utilize all the spatial location, local feature shape and orientation information of a feature based on the popular GMM framework. Experimental results show that using the additional information can obtain much better robustness, recall and convergence speed than using the location information only.

Note that the framework in this paper can be used for other GMM feature matching algorithms freely together with their regularization terms. Therefore, future works will be focused on the adaptation of our framework to other state-of-the-art algorithms.

ACKNOWLEDGMENT

T